%% file: main.tex
\documentclass[conference]{IEEEtran}
\IEEEoverridecommandlockouts
\usepackage{cite}
\usepackage{amsmath,amssymb,amsfonts}
\usepackage{algorithmic}
\usepackage{graphicx}
\usepackage{textcomp}
\usepackage{xcolor}
\def\BibTeX{{\rm B\kern-.05em{\sc i\kern-.025em b}\kern-.08em
    T\kern-.1667em\lower.7ex\hbox{E}\kern-.125emX}}

\usepackage{shortcuts}
\usepackage{amsmath,amssymb,amsfonts}
\usepackage{algorithmic}
\usepackage{multirow}	
\usepackage{enumitem}	
\usepackage{array}		
\usepackage{comment}    
\usepackage{url}        
\usepackage{makecell} 	
\usepackage{framed}		
\usepackage{xspace}	    
\usepackage{boldline}	
\usepackage{enumitem}	
\usepackage[misc]{ifsym}
\usepackage[scaled=0.85]{beramono}
\usepackage[T1]{fontenc}

\newcolumntype{H}{>{\setbox0=\hbox\bgroup}c<{\egroup}@{}}
\setlist{leftmargin=1.25em}

\newcommand\toolname[1][]{our approach\xspace}
\renewcommand{\paragraph}[1]{\vspace{3pt}\par\noindent\textbf{#1}.}

\begin{document}

\title{\textsc{Paoding}: A High-fidelity Data-free Pruning Toolkit\\for Debloating Pre-trained Neural Networks}

\author{\IEEEauthorblockN{Anonymous Authors}}

\author{\IEEEauthorblockN{Mark Huasong Meng\IEEEauthorrefmark{1}\IEEEauthorrefmark{2},
		Hao Guan\IEEEauthorrefmark{3}\IEEEauthorrefmark{4}, 
		Liuhuo Wan\IEEEauthorrefmark{3},
		Sin Gee Teo\IEEEauthorrefmark{2}, 
		Guangdong Bai\IEEEauthorrefmark{1}\IEEEauthorrefmark{3}\textsuperscript{\Letter},
		Jin Song Dong\IEEEauthorrefmark{1}}
	\IEEEauthorblockA{\IEEEauthorrefmark{1} National University of Singapore, Singapore, 
		\IEEEauthorrefmark{2} Institute for Infocomm Research (I2R), A*STAR, Singapore, \\
		\IEEEauthorrefmark{3} The University of Queensland, Australia, 
		\IEEEauthorrefmark{4} Southern University of Science and Technology, China\\
		Email: \{menghs, teo\_sin\_gee\}\@i2r.a-star.edu.sg, \{hao.guan, liuhuo.wan, g.bai\}@uq.edu.au, dcsdjs@nus.edu.sg}
}
\maketitle

\begin{abstract}
\input{sections/abstract}
\end{abstract}


\section{Introduction}
\input{sections/intro}

\section{Overview of \codename}
\input{sections/overview}


\section{Sampling Strategy}
\input{sections/approach}

\section{Evaluation}
\input{sections/experiment}

\section{Conclusion}
\input{sections/conclusion}


\bibliographystyle{IEEEtran}
\bibliography{reference}

\end{document}

%% file: sections/abstract.tex
We present \codename, a toolkit to debloat pre-trained neural network models through the lens of \emph{data-free pruning}. 
To preserve the model fidelity, \codename adopts an iterative process, which dynamically measures the effect of deleting a neuron to identify candidates that have the least impact to the output layer. 
Our evaluation shows that \codename can significantly reduce the model size, generalize on different datasets and models, and meanwhile preserve the model fidelity in terms of test accuracy and adversarial robustness.
\codename is publicly available on PyPI via \url{https://pypi.org/project/paoding-dl}. 

%% file: sections/intro.tex

Nowadays, people tend to train large neural network models containing many hidden layers for higher accuracy and better generalization.
However, the resulted models may be over-parameterized and computationally intensive, making it challenging to deploy models on devices with limited hardware capabilities, including Internet-of-Things (IoT) and mobile devices.
\emph{Neural network pruning} is a typical technique to mitigate this challenge by 
\emph{debloating} a large model, which can remove the redundant parameters without significant performance loss. 
Most existing pruning approaches~\cite{lee2018snip,li2016pruning,liu2019rethinking} are proposed based on the assumption that the pruned model has a chance to be re-trained or fine-tuned. 
However, the training data may not be available due to the protection of privacy or intellectual property.
In this work, we develop \codename to achieve \emph{data-free pruning}~\cite{srinivas2015data,tanaka2020pruning,meng2023supervised}.
Our pruning technique is designed from the model users' perspective, and 
gets rid of the re-training or fine-tuning process that entails the original dataset.
Upon receiving a pre-trained model, users can perform on-demand pruning with our approach to ease the deployment on a wide range of devices.




%% file: sections/overview.tex
Due to the absence of re-training that can fix the mis-pruned neurons\footnote{We use the term ``neurons'' to indicate the units/nodes within a layer. We also use ``hidden units'' and ``channels'' to more specifically refer to the neurons in fully-connected and convolutional layers.}, existing pruning that adopts an aggressive cut-and-re-train strategy can no longer be applied in data-free pruning.
For that reason, the design of \codename aims to preserve model fidelity, in terms of test accuracy and robustness against undesirable inputs.

\begin{figure}[t!]
	\centering
	\includegraphics[trim={0 0cm 0 0cm},clip,width=1\linewidth]{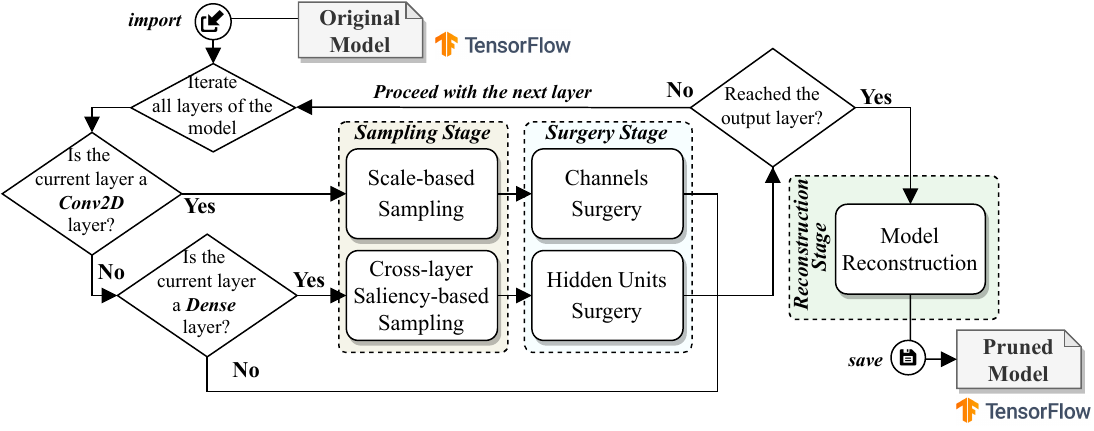}
	\centering
	\vspace{-0.65cm}
	\caption{The workflow of \textsc{Paoding} pruning}
	\vspace{-13pt}
	\label{fig:overview}
\end{figure}

Figure~\ref{fig:overview} shows the workflow of \codename.
It begins with reading a pre-trained model and loading its parameters and configurations. 
Then it iterates all the convolutional (Conv2D) and FC (dense) hidden layers and performs pruning in three stages.

\begin{itemize}[noitemsep,topsep=3pt]
\item First, it traverses all neurons of that layer and samples the pruning candidates based on specific strategies (refer to the next section) according to the layer type.
\item Given a list of candidate neurons nominated, \codename performs pruning. It  adopts a \emph{surgery} approach that deletes the candidate neurons from the model. 
\codename shrinks the Conv2D layers by deleting some \emph{channels} for every filter. For dense layers, \codename directly cuts off the entire \emph{hidden unit} including bias and all its connections to the previous and next layers.
\item Once the pruning is accomplished, \codename reconstructs the model with the pruned layers and thus, the pruned model has a shrunk structure. 

\end{itemize}

%% file: sections/approach.tex

\codename aims to minimize the impact on the model's output for the purpose of preserving the fidelity of the original pre-trained model. 
To this end, we apply different strategies for the two types of layers. 

\paragraph{Conv2D Layers} 
Our approach adopts a scale-based sampling strategy for Conv2D layers, for the purpose of prioritizing the least \emph{salient} channels during the pruning. 
To this end, we traverse all convolutional layers and calculate the $L_1$-norm of filters within each channel.
The sum of $L_1$-norm values of all the filters within a channel is assessed as the \emph{channel scale}.
Next, \codename prioritizes the channels to be pruned by sorting their channel scales in ascending order until the target of pruning on that layer has been reached.

\paragraph{Dense Layers}
We propose a \emph{pair-wise} pruning mechanism for neurons (hidden units) of dense layers, based on an assumption that \emph{a pruning approach that produces the least impact to the outputs best preserves the fidelity of the original model}.
We use $\left<a^{l}_{i},a^{l}_{j}\right>$ to denote a neuron pair (indexed as $i$ and $j$) in the $l$-th layer. 
These two neurons are supposed to be similar. 
Therefore, one of them is considered non-salient and can be replaced by the other neuron by adding all its parameters~\cite{srinivas2015data}. Next, we adopt the state-of-the-art data-free pruning approach designed for Dense layers~\cite{meng2023supervised} to find the best candidate neuron pair to be pruned.

To simply put, the \emph{impact} of pruning a neuron pair $\left<a^{l}_{i},a^{l}_{j}\right>$ is calculated based on the simulated propagation of removing the pruned neuron, i.e., $a^{l}_{i}$, starting from layer $l$ until the output layer. 
The sampling criterion of a neuron pair is jointly determined by two metrics, namely the $L_1$-norm and the Shannon's entropy of the impact. 
The former calculates the sum of change of all output nodes caused by pruning a neuron pair, and the latter measures the \textit{uniformness} of the pruning impact made on all output nodes, stipulating a higher degree of uniformness would less likely to make the pruned model mis-classify.
Neuron pairs with small values in both the metrics will be given priority for the pruning.

%% file: sections/experiment.tex

\paragraph{Implementation and Availability}
We implement \codename using Python v3.9 and evaluate it with neural network models on TensorFlow v2.4.1.
It can be accessed through PyPI with the key ``\texttt{paoding-dl}''. At the moment of writing, \codename has over 12 thousands downloads on PyPI.

\codename accepts any legitimate format of CNN and MLP models trained by TensorFlow and is compatible with various further optimization techniques such as quantization.
It also allows the user to configure the pruning target and the batch size of pruning per step.
Given a trained model, it automatically identifies the dense and Conv2D layers, prunes the neurons from them, and stops once the pruning target has been reached.

\paragraph{Evaluation Setup}
We evaluate our approach on four neural network applications.
The first model is a small MLP model for the ULB credit card fraud dataset.
The remaining three models are CNNs trained with the Brain Tumor MRI, MNIST, and CIFAR-10 datasets. 

We perform pruning progressively in multiple epochs up to 50\% of its neurons have been pruned. At each epoch, we only prune 5\% of neurons in eligible hidden layers.
After each pruning epoch, we record the file size of the pruned model, test accuracy, and the robustness\footnote{Only models \#3-4 are evaluated because their datasets have been widely studied on the subject of robustness.} against FGSM adversary. 

\begin{figure}[t]
	\centering
	\includegraphics[trim={0 0cm 0 0cm},clip,width=1\linewidth]{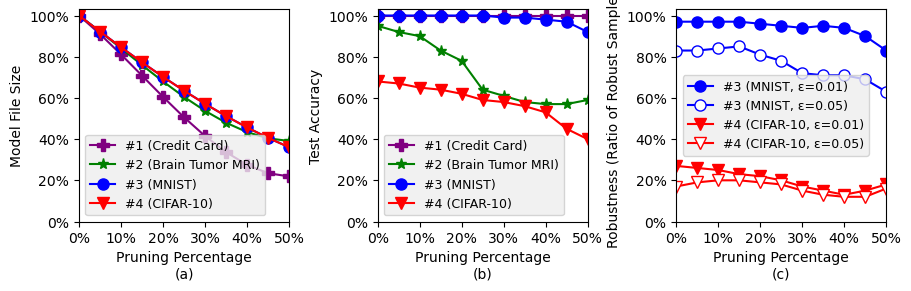}
	\vspace{-0.5cm}
	\centering
	\caption{Evaluation of model size, test accuracy, and robustness of models debloated by \codename}
	\vspace{-10pt}
	\label{fig:eval}
\end{figure}

\paragraph{Effectiveness of \codename}
\codename shows effective in debloating all four tested models. 
As shown in Figure~\ref{fig:eval} (a), a 25\% pruning can remove up to 49.4\% (model \#1) of parameters in our tested models.
By pruning another 25\% of neurons in these four models, our approach can remove 66.7\% of parameters in average (63.9\%-78.1\%), which is equivalent to 2.5x-4.5x of shrinking.

In addition to compressing model size, \codename also aims to preserve model fidelity in terms of prediction accuracy and robustness against undesirable inputs.
According to Figure~\ref{fig:eval} (b), we find that all four models show promising accuracy even after 25\% of neurons have been pruned.
This, by jointly interpreting the decay of model size, implies that we can reduce around 40\% of their parameters (36.5\%-49.4\%) without incurring a significant sacrifice of test accuracy.
We also observe less than 50\% of accuracy decay on all tested models even after pruning 50\% of their neurons.

We also assess the robustness against adversarial input perturbations (see Figure~\ref{fig:eval} (c)).
Although the tested models are not specially trained for robustness, our approach can still preserve the robustness of the pruned model and ensure the robustness declines at a controllable pace during the pruning. 
Both models maintain almost the same robustness until 20\% of its neurons have been pruned, and preserve 50\% of its original robustness even after 50\% of its parameters have been pruned.

\paragraph{Demonstration}
An online Python notebook demonstration is hosted on Google Colab. The script can be accessed through the link {\url{https://tinyurl.com/mr47vrmr}.
The demonstration begins with installing \codename through \texttt{pip} repository, followed by training a sample neural network from scratch, configuring \codename settings, and pruning the trained model.

%% file: sections/conclusion.tex

In this work, we propose \codename, a pre-trained neural network model pruning toolkit in a data-free context.
We implement our pruning as a conservative progressive process to preserve the fidelity of the model after pruning. 
Our evaluation shows that \codename can significantly shrink the model size by up to 4.5x by applying our pruning approach, at the cost of losing less than 50\% of its original accuracy and robustness on all four tested models.
Our toolkit is made publicly available and would encourage future research on exploring more optimization techniques for neural networks.